\documentclass[journal]{IEEEtran}
\IEEEoverridecommandlockouts

\newcommand\eat[1]{}

\usepackage{longtable} 
\usepackage{booktabs} 
\usepackage{cite}
\usepackage{amsmath,amssymb,amsfonts}
\usepackage{algorithmic}
\usepackage{graphicx}
\usepackage{textcomp}
\usepackage{xcolor}
\usepackage{xspace}
\usepackage[hyphens]{url}
\usepackage[hidelinks]{hyperref}
\usepackage{multirow}
\usepackage{tabularx}
\usepackage{multirow}
\usepackage{array}
\usepackage{caption}
\usepackage{longtable}
\usepackage{tikz}
\usetikzlibrary{shapes,arrows,positioning}
\usepackage{booktabs}
\usepackage{graphicx}
\usepackage{xcolor}
\usepackage{tikz}
\usetikzlibrary{shapes.geometric}

\hypersetup{breaklinks=true}
\begin{document}
\title{Image-based Deep Learning for Smart Digital Twins: a Review\\
}

\author{\IEEEauthorblockN{Md Ruman Islam\IEEEauthorrefmark{1}, Mahadevan Subramaniam\IEEEauthorrefmark{2} and
Pei-Chi Huang\IEEEauthorrefmark{3}} \\
\IEEEauthorblockA{Department of Computer Science, \\
University of Nebraska at Omaha, \\
Omaha, NE, USA\\
Email: \IEEEauthorrefmark{1}mdrumanislam,%
\IEEEauthorrefmark{2}msubramaniam,\IEEEauthorrefmark{3}phuang@unomaha.edu}
}

\maketitle

\begin{abstract}
Smart Digital twins (\textit{SDT}s) are being increasingly used to virtually replicate and predict the behaviors of complex physical systems through continual data assimilation enabling the optimization of the performance of these systems by controlling the actions of systems. Recently, deep learning (\textit{DL}) models have significantly enhanced the capabilities of \textit{SDT}s, particularly for tasks such as predictive maintenance, anomaly detection, and optimization. In many domains, including medicine, engineering, and education,  \textit{SDT}s use image data (image-based \textit{SDT}s) to observe and learn system behaviors and control their behaviors. This paper focuses on various approaches and associated challenges in developing image-based \textit{SDT}s  by continually assimilating image data from physical systems. The paper also discusses the challenges involved in designing and implementing \textit{DL} models for \textit{SDT}s, including data acquisition, processing, and interpretation. In addition, insights into the future directions and opportunities for developing new image-based \textit{DL} approaches to develop robust \textit{SDT}s are provided. This includes the potential for using generative models for data augmentation, developing multi-modal \textit{DL} models, and exploring the integration of \textit{DL} with other technologies, including 5G, edge computing, and IoT. In this paper, we describe the image-based SDTs, which enable broader adoption of the digital twin \textit{DT} paradigms across a broad spectrum of areas and the development of new methods to improve the abilities of \textit{SDT}s in replicating, predicting, and optimizing the behavior of complex systems.

\end{abstract}


\section{introduction}\label{sec:intro}
DT is a concept of enabling seamless transfer and synchronization of data from a physical entity (the physical twin) to create a virtual world (the DT) to facilitate monitoring, simulating, and optimizing the physical entities~\cite{rosen2015importance}. A DT creates a virtual replica of the physical system that runs parallel to the physical activity. This DT is an exact digital replica of the ongoing physical process. DTs are proposed to characterize various digital simulation models alongside real-time processes. However, a significant question raised here is what the differences are between a real-world system and any models of the simulation system and concerning the arguments about what constitutes a digit twin.

Based on El Saddik~\cite{el2018digital}, some of the characteristics of DTs are (i) sensors and actuators, (ii) artificial intelligence (AI) and machine learning (ML), (iii) network for communication, and (iv) representation. The sensors and actuator devices are embedded in various electrical appliances through communications to gather data sent for on-premises analysis. The data can be applied to ML to make data more convenient and dynamic. Since the system creates and analyzes large amounts of data, a major big data engine can be involved in analytic projects, especially real-time applications.

The beginning of AI and deep learning (DL) has significantly enhanced the capabilities of DTs, particularly in image-based processing. Image-based DL models have shown remarkable promise in various applications, including object detection, image classification, computer vision, and semantic segmentation \cite{YI2022552}. These models utilize DL architectures such as convolutional neural networks (CNNs), recurrent neural networks (RNNs), and other DL architectures to analyze the image data to build the accurate and efficient DT system \cite{Mukhopadhyay9781450390408}.

However, integrating DL with DTs presents significant challenges, including the requirement for extensive, high-quality datasets to train DL models, the computational complexity of DL algorithms, and the interpretability of DL models. Furthermore, it is technically challenging to synchronize physical systems and their counter DT in real-time~\cite{Jeong2022}.

Despite these challenges, the importance of image-based DL for DTs is impactful. Image-based DL models can provide detailed representations of physical systems, which helps to build accurate and realistic DT systems. These DTs models can be used in agriculture, manufacturing, and construction, where DTs can perform real-time monitoring, predictive maintenance, and process optimization. For example, we can build a smart manufacturing system in Industry 4.0 using the DTs for real-time monitoring of the machine, product quality, and production efficiency~\cite{vachalek2017digital}.

As image-based DL models are important and to the best of our knowledge, no one has worked on this topic until we write this paper. So that we can know the related image-based DL models, recent works, DT systems, data flows, and future works. The contributions of our review are summarized below:
\begin{itemize}
    \item Comprehensive review of image-based DTs and DL integration.
    \item Detailed performance comparison of various DL models in DTs.
    \item Identification of challenges and limitations in DL model implementation and optimization in DTs.
Identify the data source, including the different types of cameras and their purpose in the DT system. 
\item We highlight the importance and potential applications of image-based DL for DTs in various sectors.
    \item Proposal of future research directions for advancing image-based DTs.
\end{itemize}

The remainder of this paper is organized as follows. Section~\ref{sec:architecture} discusses the general architecture of the DTs; Section \ref{sec:data_prep} presents image-based data collection and pre-processing as an input to train the DL models; Sections~\ref{sec:intelligence} and~\ref{sec:performance} analyze the different DL based models and compare their performance; Section~\ref{sec:applications} exemplifies the recent works of DL models in the DT systems. Next, Section~\ref{sec:current} discusses the current situation and challenges of the recent work and gives some possibilities for future directions to build the image-based SDTs system. Finally, Section~\ref{sec:conclusion} describes the conclusions of review papers.
\section{Architecture for Smart Digital Twins}\label{sec:architecture}
\begin{figure}[t!]
  \centering
  \includegraphics[width=0.9\linewidth]{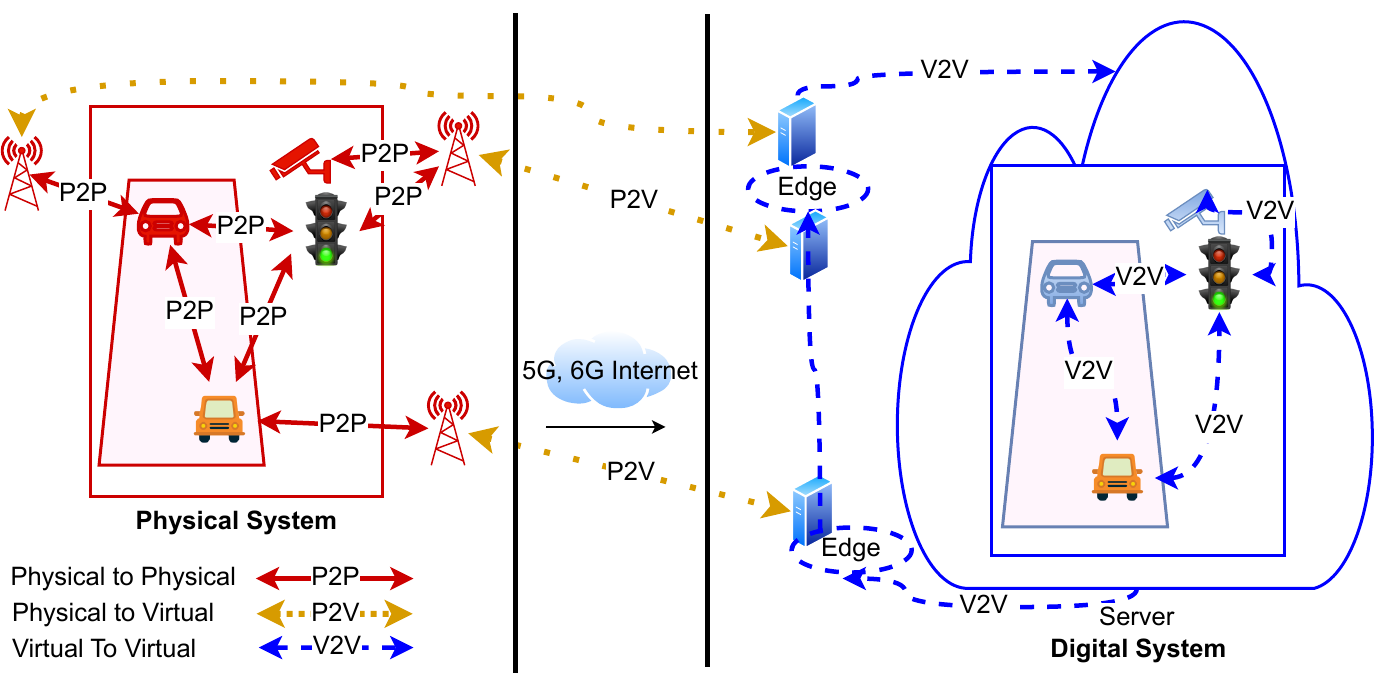}
  \caption{A general architecture of the smart (DT) system.}
  \label{fig:architecture}
\end{figure}

\begin{figure*}[htb]
  \centering
  \includegraphics[width=0.7\linewidth]{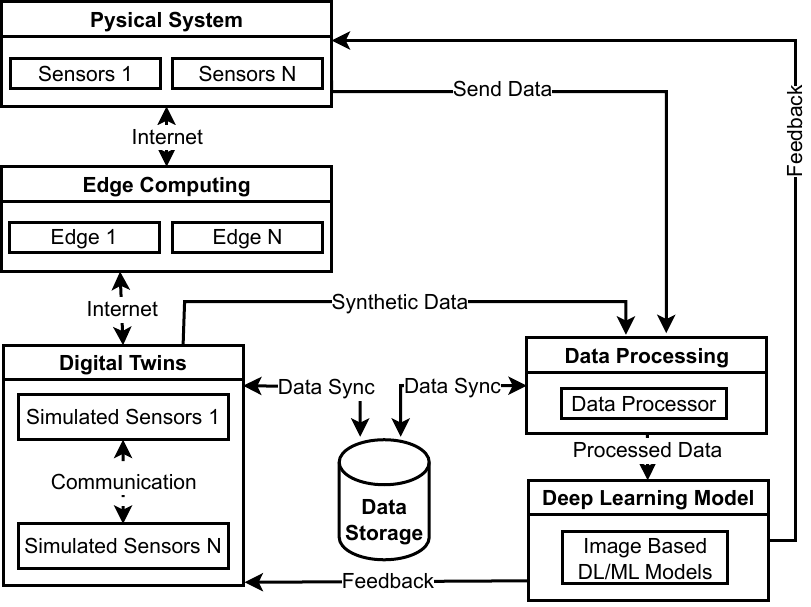}
  \caption{Dataflow and processing in the physical and digital systems.}
  \label{fig:dataflow}
\end{figure*}

The physical and digital systems play distinct roles and functions in a DT system. The physical system works as a data source of the digital system and works based on the logical decisions of digital works. On the other hand, the digital system works as a logical entity, virtual replica, or simulation of the physical system. In general, end users interact with the physical system; conversely, the digital system remains in the cloud. Sometimes, DT systems use the edge server between the two systems. The general architecture of the DT system is shown in  Figure~\ref{fig:architecture}, and we show the dataflow between these systems in Figure~\ref{fig:dataflow}. This section discusses the different components of the DT system, their purpose, and how they communicate in the DT system. In the next section, we will discuss the data acquisition and processing in the DT system. Though the DT system has two core Physical and Digital systems, we can split it into four layers. The four layers of the DT systems are the Physical System, Edge Server, Digital System, and DTN. 

In most cases, the physical system is the end user system, consisting of various sensors, actuators, devices, cameras, humidity, and temperature sensors, and physical infrastructures are used to build the physical system and to collect environmental data~\cite{Mukhopadhyay2022,Ferdousi101109CASE4999720229926529, YI2022552}. In our architecture, we consider self-driving vehicles and intelligent traffic infrastructure as examples of a DT system, where the physical system consists of surveillance cameras, self-driving vehicles, intelligent traffic signal lights, and the base stations for the network connection. The physical system is the primary source of real-time data collection. For example, cameras capture photos of the road infrastructure, send them to the digital system for analysis and synchronization, and make logical decisions for the physical system. This layer also executes the control commands generated by the digital layers and enables bidirectional communication between the two systems. 

On the other hand, digital systems work as a logical counterpart of the physical system. The digital system uses software and tools like Unity 3D, Visual Component\textregistered, and Probuilder~\cite{Mukhopadhyay9781450390408,Mukhopadhyay3489849}. It consists of computational models, algorithms, and data processing techniques that enable real-time monitoring, analysis, and prediction by collecting the information from the physical layer. Like the physical system, the digital system may also produce synthetic data to improve the system\textquotesingle s performance~\cite{Mukhopadhyay2022} ~\cite{Jeong2022}. These synthetic data are used to retrain the model to improve the accuracy of the logical decision. The digital system continuously receives the data from the physical system, makes the decisions, and sends commands to control the physical system. For instance, it receives the image data to form the physical system and also produces the synthetic data to train the ML or AI-based model to generate control commands to drive self-driving cars. After generating the commands, it sends them to the cars in a physical system. So, data flow between these two systems can be bidirectional. Together, physical and digital systems work as a framework to monitor, analyze, and control the physical system through its digital counterpart.

Edge computing processes and analyzes the data using distributed computing, storage, and network resources while communicating with the data source and the cloud. This approach effectively tackles significant challenges like energy consumption and response delays and ensures reliable wireless connections to the cloud during data transmission. Edge computing bridges the gap between the DT system\textquotesingle s physical and digital systems, known as DT edge networks (DITENs). DITENs provide secure data transmission and real-time processing, which requires heavy computational service and high bandwidth. Besides that, it can ensure security and privacy through blockchain-empowered federated learning schemes~\cite{9429703}. The purpose of the DITENs is to process and transfer data in real-time, do the computation at the edge, improve the efficiency of the computation, reduce the latency more precisely to continuously update the models, and send the commands to the physical system. 

Digital Twin Networks (DTN)  is an advanced concept that extends the idea of DT. DTN consists of various DT systems designed for the real-time exchange of information and the concurrent development of multiple physical entities alongside their digital representations. As the multiple physical and digital systems may communicate with themselves and others, the DTN is an essential part of the DT architecture. DTN is the core of a DT\textquotesingle s communication system, which establishes the communication between the different entities of the physical and digital systems. A simplified DT system relies on one-to-one communication between the physical and digital systems. In contrast, DTN maps the many-to-many communication by establishing multiple one-to-one connections. It provides the communication between the Physical to Physical (P2P), Physical to Virtual (P2V), and Virtual to Virtual (V2V). So, DTN integrates multiple DT systems to exchange information. Each type of communication has different reliability, latency, capacity, and connectivity requirements. DTN utilizes a variety of networking technologies based on the system requirements. This includes wide-area network solutions using wireless technologies such as LoRa, and cellular communication systems like 5G and 6G. Additionally, it uses a wide range of protocols, including IoT gateways and Wi-Fi access points,  wireless personal area networks (WPAN), and ZigBee to low-power wide-area network (LPWAN) technologies,  narrowband IoT (NB-IoT)~\cite{9429703}. As DTN provides communication between different components of the DT system, it is an essential part of the DT. 

The DT system consists of physical and digital systems, but edge computing and DTN ensure the efficiency of data processing and transmission in real-time access to multiple DT systems. In DT systems, the physical system works as a primary data source to train the AI/ML models in the digital system. However, the digital system may work as a secondary data source to enhance the effectiveness of the models. The digital system works as a logical counterpart of the physical system and sends the commands to the physical system for execution. The physical system works based on the logical decision of the digital system. 
\section{Images Data Acquisition and Preprocessing}\label{sec:data_prep}
Data acquisition and preprocessing play crucial roles in image-based DT systems. Different sensors in DT may capture images with varying levels of resolution, depth information, or thermal signatures, leading to diverse data characteristics. This section presents the data capabilities and limitations of the DT system.

\subsection{Sensors and Image Data Collection Techniques}

Image data collection techniques involve selecting and deploying appropriate sensors to capture visual information from the physical system. This can include 360° camera~\cite{Marai2021-mo}, thermal camera~\cite{Melesse2022-wm}, web camera~\cite{Subramanian2022-gt}, 3D Scanner~\cite{1011453571734}, depth sensors~\cite{Jeong2022}, RGB camera~\cite{YI2022552, Jeong2022} or even drones equipped with imaging capabilities. So, we can select a sensor from the wide range of sensors, but that relies on the application's unique requirements and the specific kind of visual data needed for the digital twin.

The physical and digital systems are connected through the network to communicate and exchange data in real-time~\cite{Ferdousi101109CASE4999720229926529}. Next, the physical system processes and sends the data to the digital system~\cite{1011453571734,YI2022552}. In addition, synthetic images can be generated using game engines like Unity Perception Package~\cite{unity-perception2022}, Microsoft AirSim~\cite{airsim2017fsr}, and UnrealROX~\cite{martinez2019unrealrox} in the digital system. Different DTs leverage these synthetic images to improve the training dataset and increase the models\textquotesingle performance and robustness~\cite{girshick2014rich,airsim2017fsr, martinez2019unrealrox}. Moreover, combining the real and synthetic images can generate new images to train the model effectively~\cite{Alexopoulos2020}. As synthetic data are created using the software, they are prelabeled. So we do not need extra effort to label them. The preprocess and labeling data can be stored in the Data Storage to use them in the future or to train the model~\cite{Ferdousi101109CASE4999720229926529}. ML and deep learning algorithms train the digital system on the preprocessed data. These algorithms help to analyze the data and recognize features like detecting people, objects, defects, and system surveillance~\cite{1011453571734,Jeong2022,Ferdousi101109CASE4999720229926529,YI2022552,Mukhopadhyay3489849}. Moreover, the output of the models is used to monitor and control the physical system. Finally, the predictions or outcomes made by the digital system are sent back to the physical system, where the physical works based on the feedback~\cite{Lee2020,Alexopoulos2020,1011453571734}.

In summary, smart DTusually begins with data collection from the physical system, followed by data processing and transmission to the digital system. Then, if needed, data is further processed, or synthetic images are generated in the digital system or stored in the storage. After that, different ML or deep learning models are trained on them to predict the output~\cite{9827592, 6252468}. So, an intelligent image-based DT architecture leverages image processing, deep learning, and ML techniques to create a connected system to build the smart DT. 

\subsection{Data Preprocessing Methods for Image-based Digital Twins}
Data preprocessing methods are essential to ensuring the quality and consistency of image data. Preprocessing techniques such as resizing, normalization, denoising, and image enhancement help to remove noise, improve image quality, and make the data suitable for subsequent analysis and integration into the DT. Understanding these methods helps researchers and practitioners maintain data integrity and accuracy. Data preprocessing methods are applied to the acquired image data before they are used in the DT. Preprocessing has several steps, including resizing, normalization, denoising, and image enhancement techniques to improve the quality and consistency of the data. Additionally, image registration or alignment techniques may be employed to ensure accurate spatial correspondence between the acquired images and the virtual representation in the DT.

\subsection{Challenges and Considerations in Handling Image Data for Digital Twins}
Handling image data for DTs helps raise awareness of potential obstacles and limitations. These challenges can include issues related to sensor calibration, data synchronization, storage, privacy, security, and handling variability in real-world scenarios:
\begin{itemize}
  \item There is a need to carefully calibrate and synchronize sensor data to maintain temporal alignment between the physical and virtual domains.

  \item Data storage and management must be considered, as image data can be extensive and require efficient storage and retrieval mechanisms.

  \item Privacy and security concerns need to be addressed when handling sensitive image data, ensuring compliance with relevant regulations, and safeguarding against unauthorized access.

  \item The diversity and variability of image data in real-world scenarios pose challenges in developing robust algorithms that can handle various lighting conditions, viewpoints, and object variations.
\end{itemize}
Addressing these challenges is crucial to ensure image data quality, reliability, and accuracy in the context of DTs.
\section{Image-based Intelligence Algorithms in Digital Twins}\label{sec:intelligence}
The development from a simple feature extractor~\cite{krizhevsky2017imagenet} to a taxonomy of architectures applicable across numerous computational contexts suggests that deep convolutional Neural networks (D-CNNs) can adapt to change. The artifacts of this development are grouped by commonality, and thus, one-stage and two-stage detectors~\cite{zou2023object,huang2017speed} become canonical categories for meta-architectures. Given a detector, it is considered two stages if it performs classification and regression on specific areas of interest generated by an upstream component in the network. If the detector lacks this upstream component and instead draws static box grids that exhaustively cover one or more upstream layers, then it is considered one stage. For example, the one-stage detector is the Single Shot Multibox Detector (SSD)~\cite{SSD},  which classifies and localizes objects using static anchor boxes placed at multiple feature map scales. In contrast, for two-stage detectors, Region-based Convolutional Neural Networks (R-CNN)~\cite{girshick2014rich}, widely known as the first incarnation in the hugely successful R-CNN, Fast-RCNN~\cite{girshick2015fast}, Faster-RCNN~\cite{ren2015faster} line of DCNNs, provided the first results of convolving region proposals over convolutional blocks. The subsequent improvements incrementally change parts of R-CNN to convolutional blocks until the entire network is fully convolutional and can be trained end-to-end by a given backpropagation implementation. Faster-RCNN is now considered a canonical two-stage detector and a great starting point for proposal-based detection. The reason for the architectural divergence between networks like SSD and Faster-RCNN is applicability in a given computational context.

This section primarily focuses on the various image-based DL models,  their backbone, purpose, and pros and cons, which are given in Table~\ref{tab:example}

\subsection{CNN} LeCun et al.~\cite{6795724, 726791} first propose the DL-based CNN model for classifying the handwritten digits from the image. CNN is created using different series of layers, and each layer performs a specific function. First, CNN takes the images as input and then uses the convolutional layers. The purpose of this layer is to extract or identify particular features, such as edges or corners, from the images. The result of the convolutional layer passes to the non-linear activation function, Rectified Linear Unit (ReLU), to learn complex patterns from the data. After that, the model uses a pooling layer designed to reduce the size of the feature maps by selecting the average or maximum values from each feature map segment. This layer decreases the number of parameters and combines semantically similar features, enhancing the model's robustness against noisy data. The final layer is the fully connected layer, where the output is a probability vector. The output of the final layer represents each image class.

Several improvements have been made to the vanilla version of CNN. AlexNet (2012)\cite{Krizhevsky2017}, VGG(2014)\cite{simonyan2014very}, GoogLeNet(2014)\cite{szegedy2015going} and ResNet(2015)\cite{he2016deep} are the poular among them. For example, based on the top-5 errors and the ImageNet dataset, AlexNet has an error rate of 16.4\% on eight layers, VGG has  7.3\% on 16 layers, GoogLeNet has 7.4\% on 22 layers, and ResNet has 3.6\% error on 152 layers~\cite{Pak2017}. So, an increasing number of layers decrease the error rates. 

CNN is among the most essential models in self-driving cars, computer vision, and related image-processing fields. For example, Melesse et al.~\cite{Melesse2022-wm} use the CNN-based model to process the thermal images for developing a DT system to monitor fruit quality. However, though it is widely used for image processing, it is computationally expensive because it requires a larger dataset for the training. As a result, there are better models for resource-constrained fields like IoT and embedded devices where processing power and memory are limited~\cite{HABIB20224244}. However, the primary purpose of the CNN is to classify the image instead of detecting the object. As a result, CNN somewhat suffers in detecting the object when the image has multiple objects or small objects, overlapping scenes, and low contrast with neighboring objects. Besides that, we put bounding boxes around the detected objects, and there can be multiple boxes with different sizes at different positions for a single image. Therefore, the CNN-based approach can blow up the to when the number of detected objects is enormous. In contrast, a Region-based Convolutional Neural Network (R-CNN), You Only Look Once (YOLO), or related region-based models are helpful for object detection.

\subsection{R-CNN Based Models} Girshick et al.~\cite{rcnn} propose a regional-based approach called R-CNN, which only considers ~2k regions and leverages selective search to identify the regions for the objects. The selective search first generates initial sub-segmentation, then recursively merges small and similar regions and calculates the final candidate region proposal. Then, each final region is fed into different CNN models, developed using five convolutional and two fully connected layers. As CNN models need a specific size of an image, they convert the image data for a region to a fixed pixel size of $227 \times 227$. However, as an image could have multiple objects and their size can be different, they produce regions with different sizes. So, the authors of R-CNN force them to fit into a fixed size. Here, CNN extracts a features vector of 4096 dimensions for each region from the output 2000 features regions. Finally, the SVM uses CNN outputs to classify the detected objects. They use regression to determine the bounding boxes and their positions and size in an image. They do not use all the regions. To get the correct area for a bounding box, they use a technique called Non-Max Suppression. This technique keeps the boxes based on the Intersection Over Union (IoU) scores. They keep a box if the score exceeds or equals 0.5. They suppress the boxes which are less than this threshold. There can be multiple bounding boxes for an object. They discard all the bounding boxes for an object except the one with the highest scores.  


Though the R-CNN proposes limiting the number of region proposals to 2000, They run the CNN models for each region separately, increasing the processing time to 47 seconds to process an image. As a result, it is impractical to implement in a real-time system. Besides that, they use fixed selective search algorithms for selecting the regions. Unfortunately, this algorithm has no learning capabilities like ML algorithms, so it may struggle to select the proper candidate regions. 

The same authors of R-CNN propose another model called fast R-CNN~\cite{fastrcnn} using a single CNN model for all regions, which is nine times faster than the R-CNN models. Though the Fast R-CNN model is identical to the R-CNN model,  this model feeds the whole image to the CNN models to generate the feature maps. Then, after collecting the required region proposals from the feature maps using the max pooling, they reshape the sizes to a fixed square. Finally, they use the RoI (Region of Interest) pooling layer to reshape the feature maps to pass them to the output layer. The output of RoI is that the feature vectors are passed to a softmax layer to calculate the probability values for the object's class and offset values for the bounding box.


Though Fast R-CNN performs better than R-CNN, it still has the bottleneck of selective search to compute the region proposal. Moreover, this search algorithm must be faster and learn to select the region's proposals over time. Therefore, Ren et al.~\cite{fasterrcnn}  propose an updated version of Fast-RCNN called Faster R-CNN~\cite{Faster_R-CNN} by eliminating the selective search to select the region proposal using the Region Proposal Network (RPN). Like Fast R-CNN, it also feeds the image to the CNN for the convolutional feature maps. However, instead of using the selective search to predict the region's proposals from the CNN outputs, it uses the RPN. Then, the RoI pooling layer is leveraged to reshape the region's proposals. Finally, the result of the RoI pooling layers is used to classify the objects and calculate the bounding box size and position. This algorithm is far faster than the Fast R-CNN and takes around 0.2 seconds to process an image. This model is also used to build the image base DT. For example, Wu et al.~\cite{1011453571734} use Faster-RCNN to create the DT to detect faulty surfaces from colored images. 

\subsection{YOLO Based Models} Though the best regional-based model, Faster R-CNN process time is 5fps, it is not fast enough to detect real-time objects. Besides that, regional-based object detection algorithms do not check the whole image at a time. In contrast, they check the regions with a high probability of having objects. Moreover, their computational power is high and time-consuming. So, Redmon et al.~\cite {YOLO} propose the You Only Look Once (YOLO) considerably faster than the Faster R-CNN models. The faster model can process 145fps, while the standard model with accuracy can process 45fps. 

After taking the images as input, it generates an $S \times S$ size grid over each image. This model identifies the object by checking the center of the object in a cell. There can be multiple $B$ bounding boxes for a grid cell, so a grid cell calculates the confidence scores for all the bounding boxes whose centers lie in that cell. Confidence scores are used to determine the model's confidence, whether a box has an object or not. Besides that, this score also eliminates the boxes based on the values if a cell has multiple boxes. The confidence scores of a cell for an object are equal to the Intersection Over Union (IOU) value. However, the scores will be zero if the cell is empty. Along with the confidence scores, each of the boxes calculates four mores values, which are  $(x,y)$ for the center of the cell relatives to the bounding box, $w$, and $h$ define the height and width of a cell relative to the whole image. Furthermore, they also calculate one set of conditional class probability $C = \text{Pr}(\text{Class}_i | \text{object})$ for each of the cells of a bounding box $B$. Then, they take the multiplication value of the confidence scores and conditional probabilities to determine the class-specific confidence scores for boxes. These scores determine the fitness of the objects in the bounding box. They use the  $S = 7$, $B = 2$, and $C = 20$ to evaluate their model on PASCAL VOC. Their final prediction is a tensor of size  $7 \times 7 \times 30$ because they encoded the prediction as  $S \times S \times (B \times 5 + C)$ tensor. They use a CCN model of 24 convolutional and two fully connected layers. The initial layers extract the features, and the fully connected output layers compute the probability and coordinate for an object.

As the YOLO is super fast at detecting objects, it can be used to process images in real-time. For instance, Pengnoo et al.~\cite{Pengnoo_9121234} build a DT system for detecting the object on YOLO. Though YOLO is fast for detecting objects, it still suffers from strong spatial constraints because each grid cell belongs to only one class. As a result, it struggles to predict the small objects that come into groups. 

YOLO is used in different applications where fast detection is required. Five significant versions of the YOLO are released, and among them, all of them are released by the previous authors, except the YOLOv5, released by Jocher et al.~\cite{YOLOv5}. YOLOv3, YOLOv4, and YOLOv5 are the most recent and popular versions. Besides having the full versions, they have tiny versions with lower layers and higher frame rates. YOLOv3 is built on the Darknet-53 network. This version uses the cross-entropy loss function to solve the problem of mean squared error. Besides that, it uses the softmax classifier to predict a bounding box's class. On the other hand, YOLOv3-tiny leverages the  Darknet-19, consisting of 19 convolutional layers. This one is faster than the full version. YOLOv4 has more layers than v3, and this version uses the Complete-IoU as a loss function to predict the bounding box more accurately. YOLOv5 is developed using Python instead of C using the framework PyTorch to train it faster. Like the previous version, YOLOv5 has four sub-versions: s, m, l, and x. Each has the performance variability to control the speed and the accuracy. Sozzi et al.~\cite{sozzi2022automatic} experimented on the different versions of YOLO to check their performance and accuracy. They evaluate how each version detects the grapes bunch and counts in real-time. The accuracy of different versions at mean Average Precision $(mAP)@50$ is 0.564 (v3-tiny ), 0.657 (v4-tiny),  0.775 (v5s),  0.729 (v3), 0.792 (v4 ), and  0.796 (v5x) whereas they achieve the FPS 200 for v3-tiny, 196 for v4-tiny, 61.1 for v5s, 34.7 for v3, 31.1 for v4, and 32.2 for v5x . So, v5x can predicate the objects more accurately, and v3 and v4 are faster than v5. So, if we need a faster model, we must sacrifice accuracy. As YOLO-based models are fast and have balanced accuracy, several works have been done to build the DT model using different versions of the YOLO. For example, ~\cite{Mukhopadhyay2019},~\cite{Jeong2022}, ~\cite{Mukhopadhyay9781450390408},~\cite{Mukhopadhyay3489849} build their DT model using YOLOv3 and on the other hand  Zhou et al.~\cite{9827592} build their using YOLOv5.

\subsection{MediaPipe} 
MediaPipe is a platform-independent and open-source DL-based framework in computer vision developed by Google~\cite{mediapipe}. This platform-independent framework can be set up in iOS, Android, desktop, cloud, and IoT platforms. After being built on one platform, this framework can be transferred to another platform for deployment. Besides that, it is an open-source product so that developers can modify it based on their requirements. Otherwise, they can use the prebuilt models. For example, in a DT system, Subramanian et al.~\cite{Subramanian2022-gt}  use this model to detect the face, hand, and body pose in real-time. Architecturally, this framework supports the GPU, CPU, and TPU to train the model quickly. In addition, it has many more features, including 2D and 3D object detection and tracking, face detection, hand tracking, pose detection and tracking, hair segmentation, etc. So, this is an ideal framework to build the model with less effort to give support to multiple frameworks. 

\subsection{Swin Transformer (Swin-T)}
Transformers-based deep learning is a modern technique in NLP, while Transformer Vision (ViT) learning is comparatively new in computer vision. However, ViT has several limitations, including inefficient memory access and latency. For example, sliding-window-based self-attention layers in current approaches need more efficient memory access. On the other hand, as CNNs are the backbone of computer vision, there is potential for the ViT model to create a unified model across vision and language domains. Moreover, existing self-atten-based architectures, which replace spatial convolution layers with self-attention, suffer from significantly larger actual latency than convolutional networks. However, it can increase the accuracy/FLOPs trade-offs. Currently, existing ViT models show promising results in classifying the image. However, these architectures are impractical for multi-functional networks in image analysis tasks or when dealing with high-resolution images. Because they create feature maps that lack detail, they require high computation to process high-resolution images. In contrast, Swin Transformer (Swin-T)~\cite{swin} addresses the mentioned issues and performs better.
It acquires 87.3 top-1 accuracy on ImageNet-1K and 51.1 masks AP on COCO test-dev. This model can be used for several purposes, including classification, detection, and segmentation. For example, Wu et al.~\cite{1011453571734} use the Swin-T network for feature extraction from color imaging data to build a DT system.

\subsection{3D-VGG and 3D-ResNet}
The optical flow method extracts temporal video information by calculating pixel positions' changing distance and direction between adjacent frames. However, the traditional two-stream network, which employs optical flow for feature extraction, is computationally expensive and unsuitable for human-machine interaction (HMI) applications, where real-time responsiveness is crucial.
    
To address this issue, Wang et al.~\cite{Wang2021} propose using 3D-VGG and 3D-ResNet models based on 3D Convolutional Neural Networks (3DCNN)~\cite{Ji2013-ay}. These models utilize three-dimensional convolution kernels to extract spatial and temporal features simultaneously, resulting in more efficient data analysis for HMI problems. The 3D-VGG model is an adaptation of the VGG model, initially designed for two-dimensional image classification tasks. The model retains the VGG structure but modifies the different parameters, including the input size, channels, and convolution kernels for 3D data. The 3D-ResNet model is based on the ResNet architecture, which introduces a residual block to address training difficulties in deep networks. The model uses a three-dimensional input and modifies the convolution blocks and pooling layers accordingly.
    
The authors aim to enhance HMI in DT technology by incorporating 3D-VGG and 3D-ResNet models, prioritizing real-time performance. These models analyze visual information and transform it into action or skeleton data to guide HMI. Furthermore, they generate human skeletal data from video and facilitate better collaboration in virtual environments.

\subsection{Single Shot Detector (SSD)}
The Single Shot Detector (SSD)~\cite{SSD} is a DL-based object detection model. This model aims to identify the object from the images and video frames effectively. It can detect the object by just a single pass in the neural network. SSD combines the strengths of both region proposal networks (RPNs) and convolutional neural networks (CNNs) to achieve real-time object detection. It uses a series of convolutional layers with different spatial resolutions, known as feature maps, to capture information. It uses these feature maps to determine the bounding boxes and calculate the probability of the objects of various sizes within the input image.

The SSD architecture consists of a base CNN, such as VGGNet~\cite{simonyan2014very} or ResNet, followed by additional convolutional auxiliary or detection layers. These detection layers make predictions at different spatial resolutions and scales. They use predefined anchor boxes, reference boxes of distinct aspect ratios and sizes, to generate region proposals for potential objects.

During training, SSD utilizes a combination of loss functions to optimize the model. These loss functions include confidence loss (e.g., cross-entropy loss) to measure object detection accuracy and localization loss (e.g., smooth L1 loss) to assess the precision of bounding box predictions. The advantages of SSD include its real-time processing capabilities, high detection accuracy, and the ability to handle objects of various sizes and aspect ratios effectively. Autonomous driving, surveillance systems, and object tracking systems use this approach extensively where fast and accurate object detection is required. For instance, Marai et al.~\cite{Marai2021-mo} use this model to build the DT system to detect objects for surveilling the road infrastructure.



\begin{table*}[htbp]
  \caption{A Comparison of Images Processing for Deep Learning (DL) in Digital Twin}
  \label{tab:example}
  \centering
   \begin{tabular}{|*{6}{p{2.5cm}|}}
    \hline
    
    \textbf{DL Models} & \textbf{Backbone} & \textbf{Related Works in DT} & \textbf{Purposes} & \textbf{Pros} & \textbf{Cons} \\
    \hline

    CNNs & Convolutional layers, Activation and loss function, pooling and fully connected layers   & Melesse et al.~\cite{Melesse2022-wm} \newline Alexopoulos et al.~\cite{Alexopoulos2020} \newline YI et al.~\cite{YI2022552} \newline Ferdousi et al.~\cite{Ferdousi101109CASE4999720229926529} & Image classification. Classifying thermal images of fruits in real-time & Designed specifically for image processing & Large number of parameters, prone to overfitting \\ \hline

    Faster R-CNN~\cite{Faster_R-CNN} & CNN, typically ResNet & Wu et al.~\cite{1011453571734} & Object detection & High accuracy, effective for complex scenes & Slower than YOLOv5, requires region proposals \\ \hline

    YOLO~\cite{YOLO} \newline\newline YOLOv3~\cite{yolov3} \newline\newline YOLOv5~\cite{YOLOv5} & CNN \newline\newline CSPDarknet-53 \newline\newline & Pengnoo et al.\cite{Pengnoo_9121234} \newline\newline Mukhopadhyay et al.\cite{Mukhopadhyay2019} \newline Jeong et al.~\cite{Jeong2022} \newline Mukhopadhyay et al.~\cite{Mukhopadhyay9781450390408} \newline Mukhopadhyay et al.~\cite{Mukhopadhyay3489849}  \newline\newline Zhou et al.~\cite{9827592}& Detecting objects &YOLO outperforms Faster R-CNN in speed, making it highly effective for real-time object detection. \newline\newline High accuracy and speed, real-time object detection. & Faces difficulties detecting grouped objects due to using a fixed number of grid cells, each linked to a single class. \newline\newline High computational requirements, limited accuracy on small objects and occlusions. \\ \hline

    SSD~\cite{SSD} & VGG-16, MobileNet, ResNet & Marai et al.~\cite{Marai2021-mo} & Object detection & Fast and efficient object detection for large objects. Faster than YOLO and Faster R-CNN. & Limited accuracy on small objects and difficulty handling large-scale and aspect ratio variations.\\\hline

    MediaPipe\cite{mediapipe} & Depends on the specific pipeline & Subramanian et al.~\cite{Subramanian2022-gt}  & For detecting of face, hand, and body pose in real-time.  & Real-time performance supports various ML models, easy to integrate & Limited to the models and features provided by the framework \\ \hline

    Swin-T~\cite{swin}  & Transformer & Wu et al.~\cite{1011453571734} & image classification, object detection, semantic segmentation, and others. & Efficient and performs well on many vision tasks & It struggles to handle when elements are in different sizes and requires more computing power for high-quality images. \\ \hline

    3D-VGG and 3D Resnet~\cite{Wang2021} & VGG, and RestNet &  Wang et al.~\cite{Wang2021} &  Video classification. Generating human skeletal data from video & Better performance in human action recognition tasks compared to traditional models & Requires high computing power for real-time video analysis \\ \hline
    
  \end{tabular}
\end{table*}

\section{Performance Comparison of Deep Learning Models}\label{sec:performance}
Deep learning models are increasingly popular in digital twins, image processing, and computer vision tasks like image classification, object detection, and segmentation. Various architectures and methods are used to enhance the accuracy and efficiency of these deep learning models. However, the performance of these models does not solely depend on the model\textquotesingle s architecture but also on the specific use cases they are designed for, the type of data (2D or 3D), and size of the data set, resolution of and quality of the images, batch sizes. This section will thoroughly compare these models based on their performance metric, presented in Table~\ref{tab:performance}.

\vspace{0.025in}
\noindent {\bf Image Classification}: CNNs are specifically designed to classify 2D images, and their performance is increasing rigorously with the release of the new architecture. For instance, the Top-5 error rate of AlexNet on the ImageNet dataset is 16.4\%. However, as model architectures developed, this error rate decreased with the release of VGG, GoogLeNet, and ResNet, and their error rates are 7.3\%, 6.7\%, and 3.6\%, respectively~\cite{Pak2017}. These models use the updated architecture to improve performance. For example, VGG used smaller, more consistent convolutional filters, GoogLeNet introduced the inception module, and ResNet utilized residual connections.

 
\begin{table*}[htbp]
  \caption{Performacne of Different DL Models.}
  \label{tab:performance}
  \centering
\begin{tabular}{|l|l|l|l|l|ll|l|l|}
\hline
\textbf{Model} & \textbf{Process} & \textbf{Real-Time?} & \textbf{Version} & \textbf{Dataset} & \multicolumn{2}{l|}{\textbf{Performance}} & \textbf{FPS} & \textbf{Ref.} \\ \hline
\multirow{4}{*}{CNN (1998)} & \multirow{4}{*}{2D \& 3D} & \multirow{4}{*}{Depends} & AlexNet & \multirow{4}{*}{ImageNet} & \multicolumn{1}{l|}{\multirow{4}{*}{Top-5 Error}} & 16.4\% & \multirow{8}{*}{-} & \multirow{4}{*}{\cite{Pak2017}} \\ \cline{4-4} \cline{7-7}
 &  &  & VGG &  & \multicolumn{1}{l|}{} & 7.3\% &  &  \\ \cline{4-4} \cline{7-7}
 &  &  & GoogLeNet &  & \multicolumn{1}{l|}{} & 6.7\% &  &  \\ \cline{4-4} \cline{7-7}
 &  &  & ResNet &  & \multicolumn{1}{l|}{} & 3.6\% &  &  \\ \cline{1-7} \cline{9-9} 
\multirow{4}{*}{R-CNN (2015)} & \multirow{4}{*}{2D} & \multirow{4}{*}{Usually No} & R-CNN & PASCAL VOC 2010 & \multicolumn{1}{l|}{\multirow{7}{*}{mAP}} & 53.7\% &  & \cite{rcnn} \\ \cline{4-5} \cline{7-7} \cline{9-9} 
 &  &  & Fast R-CNN & PASCAL VOC 2007+2012 & \multicolumn{1}{l|}{} & 68.4\% &  & \multirow{6}{*}{\cite{SSD}} \\ \cline{4-5} \cline{7-7}
 &  &  & \multirow{2}{*}{Faster R-CNN} & PASCAL VOC 2007+2012 & \multicolumn{1}{l|}{} & 70.4\% &  &  \\ \cline{5-5} \cline{7-7}
 &  &  &  & PASCAL VOC 2007+2012, COCO & \multicolumn{1}{l|}{} & 75.9\% &  &  \\ \cline{1-5} \cline{7-8}
\multirow{9}{*}{YOLO (2016)} & \multirow{9}{*}{2D} & \multirow{9}{*}{Yes} & Fast YOLO & PASCAL VOC 2007 & \multicolumn{1}{l|}{} & 52.7\% & 155 &  \\ \cline{4-5} \cline{7-8}
 &  &  & \multirow{2}{*}{YOLO (VGG16)} & PASCAL VOC 2007 & \multicolumn{1}{l|}{} & 66.4\% & 21 &  \\ \cline{5-5} \cline{7-8}
 &  &  &  & PASCAL VOC 2007+2012 & \multicolumn{1}{l|}{} & 57.9\% & - &  \\ \cline{4-9} 
 &  &  & V3-tiny & \multirow{6}{*}{OIDv6 + GrapeCS-ML} & \multicolumn{1}{l|}{\multirow{6}{*}{mAP@50}} & 56.4\% & 200 & \multirow{6}{*}{\cite{sozzi2022automatic}} \\ \cline{4-4} \cline{7-8}
 &  &  & V3 &  & \multicolumn{1}{l|}{} & 72.9\% & 34.7 &  \\ \cline{4-4} \cline{7-8}
 &  &  & V4-tiny &  & \multicolumn{1}{l|}{} & 65.7\% & 196 &  \\ \cline{4-4} \cline{7-8}
 &  &  & V4 &  & \multicolumn{1}{l|}{} & 79.2\% & 31.1 &  \\ \cline{4-4} \cline{7-8}
 &  &  & V5s &  & \multicolumn{1}{l|}{} & 77.5\% & 61.1 &  \\ \cline{4-4} \cline{7-8} 
 &  &  & V5x &  & \multicolumn{1}{l|}{} & 79.66\% & 32.2 &  \\ \hline
\multirow{6}{*}{SSD (2016)} & \multirow{6}{*}{2D} & \multirow{6}{*}{Yes} & \multirow{3}{*}{SSD300} & PASCAL VOC 2007+2012 & \multicolumn{1}{l|}{\multirow{6}{*}{mAP}} & 72.4\% & \multirow{2}{*}{-} & \multirow{6}{*}{\cite{SSD}} \\ \cline{5-5} \cline{7-7}
 &  &  &  & PASCAL VO C2007+2012, COCO & \multicolumn{1}{l|}{} & 77.5\% &  &  \\ \cline{5-5} \cline{7-8}
 &  &  &  & PASCAL VOC 2007 & \multicolumn{1}{l|}{} & 74.3\% & 59 &  \\ \cline{4-5} \cline{7-8}
 &  &  & \multirow{3}{*}{SSD512} & PASCAL VOC 2007+2012 & \multicolumn{1}{l|}{} & 74.9\% & \multirow{2}{*}{} &  \\ \cline{5-5} \cline{7-7}
 &  &  &  & PASCAL VOC2007+2012, COCO & \multicolumn{1}{l|}{} & 80\% &  &  \\ \cline{5-5} \cline{7-8}
 &  &  &  & PASCAL VOC2007 & \multicolumn{1}{l|}{} & 76.8\% & 22 &  \\ \hline
MediaPipe (2019) & 2D \& 3D & Yes & - & - & \multicolumn{2}{l|}{Varies by configuration} & - & - \\ \hline
\multirow{2}{*}{3D-VGG (2021)} & \multirow{2}{*}{3D} & \multirow{2}{*}{Usually No} & \multirow{2}{*}{-} & UCF-101 & \multicolumn{1}{l|}{\multirow{4}{*}{Accuracy}} & 84.6\% & \multirow{4}{*}{-} & \multirow{4}{*}{\cite{Wang2021}} \\ \cline{5-5} \cline{7-7}
 &  &  &  & HMDB-51 & \multicolumn{1}{l|}{} & 54.1\% &  &  \\ \cline{1-5} \cline{7-7}
\multirow{2}{*}{3D-ResNet (2021)} & \multirow{2}{*}{3D} & \multirow{2}{*}{Depends} & \multirow{2}{*}{-} & UCF-101 & \multicolumn{1}{l|}{} & 86.2\% &  &  \\ \cline{5-5} \cline{7-7}
 &  &  &  & HMDB-51 & \multicolumn{1}{l|}{} & 53.2\% &  &  \\ \hline
\multirow{3}{*}{Swin-T (2021)} & \multirow{3}{*}{2D \& 3D} & \multirow{3}{*}{Depends} & Classificaiton & ImageNet-1K & \multicolumn{1}{l|}{Op-1 accuracy} & 87.3\% & 42.1 & \multirow{3}{*}{\cite{swin}} \\ \cline{4-8}
 &  &  & \multirow{2}{*}{Object detection} & \multirow{2}{*}{COCO} & \multicolumn{1}{l|}{Box AP} & 58.7\% & \multirow{2}{*}{-} &  \\ \cline{6-7}
 &  &  &  &  & \multicolumn{1}{l|}{Mask AP} & 51.1\% &  &  \\ \hline
\end{tabular}
\end{table*}

\noindent {\bf Object Detection}: R-CNN-based models focus on object detection and segmentation rather than focusing on image classification like CNN. The original R-CNN model reaches a mAP of 53.7\% on the PASCAL VOC 2010 dataset, which is improved to 75.9\% by the Faster R-CNN model on a combined PASCAL VOC and COCO dataset \cite{rcnn, SSD}. However, the ability of these models to process the data in real time depends on their computational power.

\noindent {\bf Real-time Object Detection}: Though the R-CNN-based models are designed for 2D object detection, YOLO and SSD models bring groundbreaking performance. YOLO presents a novel approach using a single neural network to detect the object. There are different versions of YOLO currently available to detect the object. The Fast YOLO model achieved an mAP of 52.7\% on the PASCAL VOC 2007 dataset, while the more advanced YOLO V4 model achieved an mAP of 79.2\%, highlighting the continuous improvement in real-time object detection \cite{sozzi2022automatic, SSD}. SSD models, such as the SSD512, have also shown significant performance, reaching an exceptional mAP of 80\% on combined PASCAL VOC 2007 + 2012 and COCO datasets \cite{SSD}. 

\noindent {\bf 3D Data Processing}: 3D-VGG and 3D-ResNet have shown promising performance in processing 3D data for detecting shapes of the object. They are specifically designed for processing 3D data. 3D-ResNet and 3D-VGG achieved an accuracy of 86.2\% and 84.6\% on the UCF-101 dataset showing its effectiveness in 3D object recognition tasks \cite{Wang2021}. 3D-ResNet performs better than the  3D-VGG for 3D objects detections. 

\noindent {\bf Mixed Use Cases}: The purpose of MediaPipe and Swin-T are mixed. For example, MediPipe is designed for real-time hand tracking, pose estimation, and face detection from both 2D and 3D data. On the other hand, Swin-T is designed for object detection, classification, and semantic segmentation in both 2D and 3D. The performance of the MediaPipe depends on its implementation, and we did not find any direct measurement matrix for it. However, the benefit of media pipe is that it is platforms independent and compatible with mobile devices. Swin-T achieved an Op-1 accuracy of 87.3\% on the ImageNet-1K dataset for image classification and a Box AP of 58.7\% on the COCO dataset for object detection, demonstrating the model\textquotesingle s performance across different tasks \cite{swin}. 

\noindent {\bf Role of Dataset Size in Model Performance}: The mAP of the models heavily depends on the data set. In general, models are performed better on large datasets. For instance, 
 Faster R-CNN demonstrates better performance on larger datasets. For example, on the PASCAL VOC 2007 + 2012 dataset, Faster R-CNN achieves an mAP of 70.4\%, which increases to 75.9\% on combined larger datasets, PASCAL VOC 2007 + 2012 and COCO. Similarly, SSD also improved the performance when trained on a larger dataset. SSD300, evaluated on the PASCAL VOC 2007 + 2012 dataset, achieves an mAP of 72.4\%. However, the performance increased to 77.5\% when trained on large datasets, PASCAL VOC 2007 + 2012 and COCO \cite{SSD}. So, performance depends on the dataset, and a more extensive dataset generally improves the performance. 

\noindent {\bf Balancing FPS and Accuracy or Precision in Real-time Applications}: Frames per Second (FPS) is an important performance matrix for image-based DL models. Higher FPS is needed for real-time applications such as video surveillance, autonomous vehicles, and other real-time monitoring systems. However, high FPS comes at the cost of lower accuracy in model performance. YOLO models, specifically Fast YOLO and YOLO V3-tiny, achieve up to 155 and 200 FPS, respectively. 
Nevertheless, Fast YOLO achieves a lower performance value in terms of precision, with a mAP of 52.7\% on the PASCAL VOC 2007 dataset. Similarly, YOLO V3-tiny, despite its high FPS, achieves a mAP of 56.4\% on the Open Image Dataset v6 (OIDv6) and GrapeCS-ML dataset. SSD models also offer high FPS, with SSD300 achieving 59 FPS and SSD512 achieving 22 FPS on the PASCAL VOC 2007 dataset. These models balance speed and accuracy, with SSD300 achieving a mAP of 74.3\% and SSD512 performing a mAP of 76.8\% on the same dataset. Therefore, when considering model performance based on FPS, it is essential to analyze the trade-off between speed and accuracy. High FPS is desirable for real-time applications, but we also need to consider precision or accuracy to ensure the application\textquotesingle s performance. 

There is no one size fits all DL model that can be used for any image processing task. Each model has advantages and disadvantages, depending on the task and dataset. Each of the models is designed to serve a specific purpose. CNN-based models are designed for classification, R-CNN is designed for object detection, YOLO and SSD are designed for faster object detection, 3D VGG, and 3D-RestNet are designed for video processing, and Swin-T and MediaPipe are designed for multipurpose. However, their performance depends on the size of the dataset and batch, resolution, accuracy, and FPS.
\section{Recent Advancements in Image-Based Deep Learning Applications for Digital Twins}
\label{sec:applications}

Image-based DTs use several DL models and different types of cameras and sensors as the data source. This section overviews the recent applications of different DL models and devices to build DT models. Detailed overview of the recent works are shared  table~\ref{tab:recentworks}. 

CNN models are used for different applications and build the DT. Ferdousi et al.~\cite{Ferdousi101109CASE4999720229926529} use different types of sensors, cameras, and CNN models to detect defection in railways. Similarly, YI et al. \cite{YI2022552} use the CNN for human-robot interaction and use several types of cameras, including Color Cameras, IR Camera, and Kinect v2.0 Camera as a data source. As previous authors, Melesse et al. \cite{Melesse2022-wm} also use the CNN model to monitor the fruit quality, utilizing thermal cameras.

\begin{table*}[htbp]
  \caption{Recent Works Utilizing DL Models and Imaging Sensors in Digital Twins}
  \label{tab:recentworks}
  \centering
   \begin{tabular}{|*{5}{p{3.05cm}|}}
    \hline
    
    \textbf{Ref} & \textbf{Application} & \textbf{Sensors} & \textbf{ML Model} & \textbf{DT Objective} \\
    \hline
    Ferdousi et al. \cite{Ferdousi101109CASE4999720229926529} & Defect Detection in Rail & Humidity, Positioning, Camera, Stereo Video Recorder, Radar, Drone, LIDAR & CNN & Defect detection in railway \\ \hline

    YI et al. \cite{YI2022552} & Human-Robot Interaction & Color Camera, IR Projector, IR Camera, Kinect v2.0 Camera & CNN & Control the robot \\ \hline

    Melesse et al. \cite{Melesse2022-wm} & Fruit Quality Monitoring & Thermal Camera & CNN & Notify end users \\ \hline

   Alexopoulos et al. \cite{Alexopoulos2020} & AI in Manufacturing & Camera & Inception-v3 CNN model, TensorFlow & Automatic data labeling \\ \hline

   Wu et al. \cite{1011453571734} & Defect Detection in Manufacturing & 2D Camera and 3D Scanner &  Faster-RCNN \cite{Faster_R-CNN}, and Swin-T \cite{swin} based deep learning model & Detect defected product, send notification \\ \hline

    Pengnoo et al. \cite{Pengnoo_9121234} & Networking & Two Cameras, Base Station, Reflector & Python simulation, YOLO & Control antenna and reflector \\ \hline

    Mukhopadhyay et al. \cite{Mukhopadhyay2022} & Workplace Safety using VR & Camera, Temperature and Humidity Sensors & CNN, YOLOv3 & Monitor social distance \\ \hline

    Mukhopadhyay et al. \cite{Mukhopadhyay3489849} & Workplace Safety & Camera & YOLOv3 & Monitor the postures distance in DT \\ \hline

    Jeong et al. \cite{Jeong2022} & Tool Damage Monitoring & Depth Camera & 2D RGB images from synthetic depth maps, YOLOv3 & Monitor tool breakage \\ \hline

    Mukhopadhyay et al. \cite{Mukhopadhyay9781450390408} & Image Processing & Camera & YOLOv3 & Improving object detection performance using synthetic data \\ \hline

    Zhou et al. \cite{9827592} & Construction Engineering & Monocular Camera & YOLOv5, MLP, Camera-BIM algorithm & Operations \& Maintenance (O\&M) \\ \hline

    Subramanian et al. \cite{Subramanian2022-gt} & Healthcare & Web Camera & MediaPipe, ML algorithms for classification & Patient Emotions detection \\ \hline

    Wang et al. \cite{Wang2021} & Human-Machine Interaction & Camera & 3D-VGG and 3D Resnet & Skeletal data from videos and real-time analysis with 3D models. \\ \hline

    Marai et al. \cite{Marai2021-mo} & Roads Infrastructure Surveillance & 360° Camera, GPS, Temperature, Humidity, Air Quality Sensors & SSD, Face\_recognition Python library, Google AI Vision API & Data stored locally, sent to OpenStack servers, enabling road monitoring. \\ \hline

  \end{tabular}
\end{table*}

 Instead of just using the CNN models, several image bases DT are built utilizing multiple models. For example, Mukhopadhyay et al. \cite{Mukhopadhyay2022} and \cite{Mukhopadhyay3489849} build the DT models using the CNN and YOLOv3 models to ensure workplace safety. For collecting the data, besides using the cameras, they use different types of sensors, including temperature and humidity sensors. 

Several versions of the CNN models are available, and Inception-v3 is one of them. Alexopoulos et al.~\cite{Alexopoulos2020} use this version to label manufacturing data automatically. Wu et al. \cite{1011453571734} build the DT model for defect detection in manufacturing using Faster-RCNN and Swin-T-based models. They use both 2D and 3D cameras for collecting the image data. 

YOLO-based models are popular for real-time object detection. Pengnoo et al.~\cite{Pengnoo_9121234} use the YOLO model in networking using two cameras, a base station, and a reflector to control the antenna and reflector. Besides the vanilla version of YOLO, several improved versions have been proposed. For example, Jeong et al. \cite{Jeong2022} use the YOLOv3 model and depth cameras for tool damage monitoring. On the other hand, Zhou et al.~\cite{9827592} use the YOLOv5 model and monocular camera for creating the Building Information Modelling/Model (BIM) based DT.

Besides the mentioned models, several specialized DL models are available to build the image-based DT. Subramanian et al. \cite{Subramanian2022-gt} use the web camera to capture the data and MediaPipe to build the DT in healthcare. Wang et al. \cite{Wang2021} propose the 3D-VGG and 3D Resnet models for processing 3D image data and validate their models by building the DT for human-machine interaction. Marai et al.~\cite{Marai2021-mo} use the SSD models and other libraries for road infrastructure surveillance, using a 360° camera and other environmental sensors.

There are several types of cameras, and DL models are available to build the image-based DT for different applications. The purpose and the use case of different types of cameras are specific. Moreover, applying the mentioned DT models varies from manufacturing and healthcare to infrastructure monitoring and human-machine interaction.
\section{Current Challenges, Situations, and Research Directions}\label{sec:current}

In the current situation, DT is gaining popularity in smart manufacturing, healthcare, energy, transportation, and smart cities. Organizations leverage DTs to optimize operations, improve maintenance, enhance product development, and enable data-driven decision-making. Also, DTs benefit from technological advancements such as IoT, cloud computing, and  AI. These technologies provide the infrastructure and tools for creating, managing, and analyzing DTs on a large scale. Moreover, DTs are increasingly integrated with physical systems, creating cyber-physical systems (CPS). This integration allows for real-time monitoring, control, and optimization of physical systems using their virtual counterparts.

\subsection{Challenges}

In DT, ML-based approaches offer flexibility but lack interpretability, while model-based verification provides rigorous analysis but may suffer from model complexity. Testing with real-world data allows for practical validation but faces data quality and coverage challenges. Hybrid approaches combining different methods can offer more comprehensive validation but introduce additional complexity. Uncertainty and trustworthiness assessment is crucial in understanding the reliability and risks associated with DT predictions. Addressing the challenges and limitations of these methods is essential to ensure the accuracy, reliability, and trustworthiness of DTs in practical applications.

\vspace{0.025in}
\noindent {\bf Machine Learning-based Approaches}: DL algorithms are trained on large datasets to learn patterns and make predictions within the DT environment. The advantages are: (i) Can handle complex and nonlinear relationships within data; (ii) Provide flexibility and adaptability to changing conditions. However, the problems are (i) lack of interpretability: DL models often lack transparency, making it challenging to understand the reasoning behind their predictions; (ii) Overfitting: Models may become too specific to training data, leading to poor generalization on unseen data; (iii) Data requirements: DL models require extensive labeled training data, which is costly to collect and process.

\vspace{0.025in}
\noindent {\bf Model-based Verification}: Verification techniques involve using mathematical models and simulations to validate the behavior of the DT against expected outcomes. The advantages are (i) Well-established mathematical techniques for verification and validation and (ii) Allowing for rigorous analysis and verification of system behavior. However, the problems are (i) model complexity: Developing accurate models for complex systems can be challenging and time-consuming; (ii) Model uncertainty: Models may not capture all aspects of system behavior accurately, which creates dissimilarities between the DT and the physical system; (iii) Calibration and validation: Ensuring that the model accurately represents the physical system requires extensive calibration and validation efforts.

\vspace{0.025in}
\noindent {\bf Testing and Validation with Real-world Data}: Real-world data collected from the physical system is used to validate the DT\textquotesingle s behavior and predictions. The advantages are to (i) provide a direct comparison between the DT and the physical system and (ii) Offer a practical validation approach by leveraging real-world data. The problems are (i) data quality and availability: Ensuring high-quality data collection and having access to diverse and representative datasets can be challenging; (ii) Limited coverage: Real-world data may not cover all possible scenarios and variations, limiting the comprehensiveness of validation; and (iii) Data synchronization: Maintaining synchronization between the DT and the physical system data can be complex, especially in real-time scenarios.

\subsection{Research Directions}
The current situation shows a growing adoption of DTs across industries, fueled by technological advancements and physical systems integration. The research directions indicate an expanding scope as follows: 
\begin{itemize}
    \item \textbf{AI and Machine Learning Integration:} The integration of AI and ML techniques with DTs is expected to accelerate. AI algorithms can enhance DTs\textquotesingle  intelligence and predictive capabilities, enabling real-time autonomous decision-making, anomaly detection, and optimization.
    \item \textbf{Real-Time and Edge Computing:} DTs are evolving to support real-time monitoring and decision-making. Edge computing technologies enable data processing and analysis closer to the physical systems, reducing latency and enabling faster responses and actions based on real-time insights.
    \item \textbf{New DTN Research Topics:} DTN is a comparatively new idea, and only a few of the research ideas have been shared by Wu et al.~\cite{9429703}. However, many research ideas still need to address, so researchers should focus on that. 
    \item \textbf{Camera Choice for Digital Twins:} Choosing the right camera is crucial for image-based DT. Several types of camera sensors are available, including IR cameras, color cameras, depth cameras, 360° cameras, monocular cameras, and web cameras to build the DT. Besides that, synthetic image is also a great source to improve the performance of DT. Choosing the correct camera or image source type is essential to build the DT.
    \item \textbf{Addressing Climate Impact on Thermal Imaging:} Melesse et al.~\cite{Melesse2022-wm} worked to detect the defect in the fruit using the thermal camera. But thermal images are effect by environmental factors such as temperature and humidity, so in the future, we should focus on building image-based DT systems using DL algorithms that can deal with environmental changes or creating a DT system that will hardly be affected by the environment\textquotesingle s variation.
    \item \textbf{Enabling Emotion Detection with Multimodal Data:} Addressing climate impact on thermal imaging in the future, we may focus on building the fusion-based DT system to improve performance. For instance, Subramanian et al.~\cite{Subramanian2022-gt} worked to detect emotions from facial expressions. However, incorporating speech recognition along with facial or body expressions may improve emotion detection performance.
    \item  \textbf{Speed-Accuracy Trade-off:} The balance between speed (FPS) and accuracy is crucial in real-time applications. However, this trade-off is not always optimized in current models. Future research should focus on creating models that maintain high speed without compromising precision/accuracy.
    \item \textbf{DL on Small Datasets:} Most DL models improved their performance using more extensive datasets. However, building and labeling a larger data set is time-consuming and sometimes impractical. Researchers should focus on enhancing DL model performance when only smaller datasets are available. 
\end{itemize}
\section{Conclusions}\label{sec:conclusion}
SDTs have emerged as valuable systems for replicating and predicting the behaviors of complex physical systems to optimize their performance. DL models have shown significant promise in enhancing the performance of SDTs, especially for maintenance, anomaly detection, and optimization. This paper provided a comprehensive overview of the development of image-based SDTs. We discussed the challenges of designing and implementing DL models for SDTs, including data acquisition, processing, and interpretation. Furthermore, the paper emphasized future directions and opportunities for developing image-based SDTs, such as utilizing generative models for data augmentation, exploring multi-modal DL approaches, and integrating them with emerging technologies like 5G, edge computing, and IoT. By leveraging the insights shared in this paper, a more comprehensive range of industries can adopt DT paradigms and develop new methods to improve the capabilities of SDTs in replicating, predicting, and optimizing complex system behaviors.


\bibliographystyle{IEEEtran}
\bibliography{bibliography}

\begin{thebibliography}{10}
\providecommand{\url}[1]{#1}
\csname url@samestyle\endcsname
\providecommand{\newblock}{\relax}
\providecommand{\bibinfo}[2]{#2}
\providecommand{\BIBentrySTDinterwordspacing}{\spaceskip=0pt\relax}
\providecommand{\BIBentryALTinterwordstretchfactor}{4}
\providecommand{\BIBentryALTinterwordspacing}{\spaceskip=\fontdimen2\font plus
\BIBentryALTinterwordstretchfactor\fontdimen3\font minus \fontdimen4\font\relax}
\providecommand{\BIBforeignlanguage}[2]{{%
\expandafter\ifx\csname l@#1\endcsname\relax
\typeout{** WARNING: IEEEtran.bst: No hyphenation pattern has been}%
\typeout{** loaded for the language `#1'. Using the pattern for}%
\typeout{** the default language instead.}%
\else
\language=\csname l@#1\endcsname
\fi
#2}}
\providecommand{\BIBdecl}{\relax}
\BIBdecl

\bibitem{rosen2015importance}
R.~Rosen, G.~Von~Wichert, G.~Lo, and K.~D. Bettenhausen, ``About the importance of autonomy and digital twins for the future of manufacturing,'' \emph{Ifac-Papersonline}, vol.~48, no.~3, pp. 567--572, 2015.

\bibitem{el2018digital}
A.~El~Saddik, ``Digital twins: The convergence of multimedia technologies,'' \emph{IEEE multimedia}, vol.~25, no.~2, pp. 87--92, 2018.

\bibitem{YI2022552}
\BIBentryALTinterwordspacing
S.~Yi, S.~Liu, X.~Xu, X.~V. Wang, S.~Yan, and L.~Wang, ``A vision-based human-robot collaborative system for digital twin,'' \emph{Procedia CIRP}, vol. 107, pp. 552--557, 2022, leading manufacturing systems transformation – Proceedings of the 55th CIRP Conference on Manufacturing Systems 2022. [Online]. Available: \url{https://www.sciencedirect.com/science/article/pii/S2212827122003080}
\BIBentrySTDinterwordspacing

\bibitem{Mukhopadhyay9781450390408}
\BIBentryALTinterwordspacing
A.~Mukhopadhyay, G.~Rajshekar~Reddy, I.~Mukherjee, G.~Kumar~Gopa, A.~Pena-Rios, and P.~Biswas, ``Generating synthetic data for deep learning using vr digital twin,'' in \emph{Proceedings of the 2021 5th International Conference on Cloud and Big Data Computing}, ser. ICCBDC '21.\hskip 1em plus 0.5em minus 0.4em\relax New York, NY, USA: Association for Computing Machinery, 2021, p. 52–56. [Online]. Available: \url{https://doi-org.leo.lib.unomaha.edu/10.1145/3481646.3481655}
\BIBentrySTDinterwordspacing

\bibitem{Jeong2022}
S.~Jeong, H.~Kim, J.~Lee, S.-Y. Park, and S.-H. Ahn, ``Digital twin-based cutting tool breakage detection model using synthetic depth map and deep learning,'' in \emph{Proc. of the 9th Intl. Conf. of Asian Society for Precision Engg. and Nanotechnology (ASPEN 2022) 15–18 November 2022, Singapore}, N.~M.~L. Sharon and A.~S. Kumar, Eds.\hskip 1em plus 0.5em minus 0.4em\relax CRC Press, 2022.

\bibitem{vachalek2017digital}
J.~Vach{\'a}lek, L.~Bartalsk{\`y}, O.~Rovn{\`y}, D.~{\v{S}}i{\v{s}}mi{\v{s}}ov{\'a}, M.~Morh{\'a}{\v{c}}, and M.~Lok{\v{s}}{\'\i}k, ``The digital twin of an industrial production line within the industry 4.0 concept,'' in \emph{2017 21st international conference on process control (PC)}.\hskip 1em plus 0.5em minus 0.4em\relax IEEE, 2017, pp. 258--262.

\bibitem{Mukhopadhyay2022}
A.~Mukhopadhyay, G.~R. Reddy, K.~S. Saluja, S.~Ghosh, A.~Pe{\~n}a-Rios, G.~Gopal, and P.~Biswas, ``Virtual-reality-based digital twin of office spaces with social distance measurement feature,'' \emph{Virtual Reality \& Intelligent Hardware}, vol.~4, no.~1, pp. 55--75, 2022.

\bibitem{Ferdousi101109CASE4999720229926529}
\BIBentryALTinterwordspacing
R.~Ferdousi, F.~Laamarti, C.~Yang, and A.~El~Saddik, ``Railtwin: A digital twin framework for railway,'' in \emph{2022 IEEE 18th International Conference on Automation Science and Engineering (CASE)}.\hskip 1em plus 0.5em minus 0.4em\relax IEEE Press, 2022, p. 1767–1772. [Online]. Available: \url{https://doi.org/10.1109/CASE49997.2022.9926529}
\BIBentrySTDinterwordspacing

\bibitem{Mukhopadhyay3489849}
\BIBentryALTinterwordspacing
A.~Mukhopadhyay, G.~S.~R. Reddy, S.~Ghosh, M.~L~R~D, and P.~Biswas, ``Validating social distancing through deep learning and vr-based digital twins,'' in \emph{Proceedings of the 27th ACM Symposium on Virtual Reality Software and Technology}, ser. VRST '21.\hskip 1em plus 0.5em minus 0.4em\relax New York, NY, USA: Association for Computing Machinery, 2021. [Online]. Available: \url{https://doi-org.leo.lib.unomaha.edu/10.1145/3489849.3489959}
\BIBentrySTDinterwordspacing

\bibitem{9429703}
Y.~Wu, K.~Zhang, and Y.~Zhang, ``Digital twin networks: A survey,'' \emph{IEEE Internet of Things Journal}, vol.~8, no.~18, pp. 13\,789--13\,804, 2021.

\bibitem{Marai2021-mo}
O.~E. Marai, T.~Taleb, and J.~Song, ``Roads infrastructure digital twin: A step toward smarter cities realization,'' \emph{IEEE Network}, vol.~35, no.~2, pp. 136--143, 2021.

\bibitem{Melesse2022-wm}
T.~Y. Melesse, M.~Bollo, V.~D. Pasquale, F.~Centro, and S.~Riemma, ``\BIBforeignlanguage{en}{Machine learning-based digital twin for monitoring fruit quality evolution},'' \emph{\BIBforeignlanguage{en}{Procedia Comput. Sci.}}, vol. 200, pp. 13--20, 2022.

\bibitem{Subramanian2022-gt}
B.~Subramanian, J.~Kim, M.~Maray, and A.~Paul, ``Digital twin model: A real-time emotion recognition system for personalized healthcare,'' \emph{IEEE Access}, vol.~10, pp. 81\,155--81\,165, 2022.

\bibitem{1011453571734}
\BIBentryALTinterwordspacing
Y.~Wu, H.~Cao, G.~Yang, T.~Lu, and S.~Wan, ``Digital twin of intelligent small surface defect detection with cyber-manufacturing systems,'' \emph{ACM Trans. Internet Technol.}, nov 2022, just Accepted. [Online]. Available: \url{https://doi.org/10.1145/3571734}
\BIBentrySTDinterwordspacing

\bibitem{unity-perception2022}
{Unity Technologies}, ``Unity {P}erception package,'' \url{https://github.com/Unity-Technologies/com.unity.perception}, 2020.

\bibitem{airsim2017fsr}
\BIBentryALTinterwordspacing
S.~Shah, D.~Dey, C.~Lovett, and A.~Kapoor, ``Airsim: High-fidelity visual and physical simulation for autonomous vehicles,'' in \emph{Field and Service Robotics}, 2017. [Online]. Available: \url{https://arxiv.org/abs/1705.05065}
\BIBentrySTDinterwordspacing

\bibitem{martinez2019unrealrox}
P.~Martinez-Gonzalez, S.~Oprea, A.~Garcia-Garcia, A.~Jover-Alvarez, S.~Orts-Escolano, and J.~Garcia-Rodriguez, ``Unrealrox: An extremely photorealistic virtual reality environment for robotics simulations and synthetic data generation,'' \emph{Virtual Reality}, pp. 1--18, 2019, [Online; accessed 6-April-2023].

\bibitem{girshick2014rich}
R.~Girshick, J.~Donahue, T.~Darrell, and J.~Malik, ``Rich feature hierarchies for accurate object detection and semantic segmentation,'' in \emph{Proceedings of the IEEE conference on computer vision and pattern recognition}, 2014, pp. 580--587.

\bibitem{Alexopoulos2020}
\BIBentryALTinterwordspacing
K.~Alexopoulos, N.~Nikolakis, and G.~Chryssolouris, ``Digital twin-driven supervised machine learning for the development of artificial intelligence applications in manufacturing,'' \emph{International Journal of Computer Integrated Manufacturing}, vol.~33, no.~5, pp. 429--439, Apr. 2020. [Online]. Available: \url{https://doi.org/10.1080/0951192x.2020.1747642}
\BIBentrySTDinterwordspacing

\bibitem{Lee2020}
\BIBentryALTinterwordspacing
J.~Lee, M.~Azamfar, J.~Singh, and S.~Siahpour, ``Integration of digital twin and deep learning in cyber-physical systems: towards smart manufacturing,'' \emph{{IET} Collaborative Intelligent Manufacturing}, vol.~2, no.~1, pp. 34--36, Mar. 2020. [Online]. Available: \url{https://doi.org/10.1049/iet-cim.2020.0009}
\BIBentrySTDinterwordspacing

\bibitem{9827592}
X.~Zhou, K.~Sun, J.~Wang, J.~Zhao, C.~Feng, Y.~Yang, and W.~Zhou, ``Computer vision enabled building digital twin using building information model,'' \emph{IEEE Transactions on Industrial Informatics}, vol.~19, no.~3, pp. 2684--2692, 2023.

\bibitem{6252468}
J.~Masci, U.~Meier, D.~Ciresan, J.~Schmidhuber, and G.~Fricout, ``Steel defect classification with max-pooling convolutional neural networks,'' in \emph{The 2012 International Joint Conference on Neural Networks (IJCNN)}, 2012, pp. 1--6.

\bibitem{krizhevsky2017imagenet}
A.~Krizhevsky, I.~Sutskever, and G.~E. Hinton, ``Imagenet classification with deep convolutional neural networks,'' \emph{Communications of the ACM}, vol.~60, no.~6, pp. 84--90, 2017.

\bibitem{zou2023object}
Z.~Zou, K.~Chen, Z.~Shi, Y.~Guo, and J.~Ye, ``Object detection in 20 years: A survey,'' \emph{Proceedings of the IEEE}, 2023.

\bibitem{huang2017speed}
J.~Huang, V.~Rathod, C.~Sun, M.~Zhu, A.~Korattikara, A.~Fathi, I.~Fischer, Z.~Wojna, Y.~Song, S.~Guadarrama \emph{et~al.}, ``Speed/accuracy trade-offs for modern convolutional object detectors,'' in \emph{Proceedings of the IEEE conference on computer vision and pattern recognition}, 2017, pp. 7310--7311.

\bibitem{SSD}
\BIBentryALTinterwordspacing
W.~Liu, D.~Anguelov, D.~Erhan, C.~Szegedy, S.~Reed, C.-Y. Fu, and A.~C. Berg, ``{SSD}: Single shot {MultiBox} detector,'' in \emph{Computer Vision {\textendash} {ECCV} 2016}.\hskip 1em plus 0.5em minus 0.4em\relax Springer International Publishing, 2016, pp. 21--37. [Online]. Available: \url{https://doi.org/10.1007/978-3-319-46448-0\_2}
\BIBentrySTDinterwordspacing

\bibitem{girshick2015fast}
R.~Girshick, ``Fast r-cnn,'' in \emph{Proceedings of the IEEE international conference on computer vision}, 2015, pp. 1440--1448.

\bibitem{ren2015faster}
S.~Ren, K.~He, R.~Girshick, and J.~Sun, ``Faster r-cnn: Towards real-time object detection with region proposal networks,'' \emph{Advances in neural information processing systems}, vol.~28, 2015.

\bibitem{6795724}
Y.~LeCun, B.~Boser, J.~S. Denker, D.~Henderson, R.~E. Howard, W.~Hubbard, and L.~D. Jackel, ``Backpropagation applied to handwritten zip code recognition,'' \emph{Neural Computation}, vol.~1, no.~4, pp. 541--551, 1989.

\bibitem{726791}
Y.~Lecun, L.~Bottou, Y.~Bengio, and P.~Haffner, ``Gradient-based learning applied to document recognition,'' \emph{Proceedings of the IEEE}, vol.~86, no.~11, pp. 2278--2324, 1998.

\bibitem{Krizhevsky2017}
\BIBentryALTinterwordspacing
A.~Krizhevsky, I.~Sutskever, and G.~E. Hinton, ``{ImageNet} classification with deep convolutional neural networks,'' \emph{Communications of the {ACM}}, vol.~60, no.~6, pp. 84--90, May 2017. [Online]. Available: \url{https://doi.org/10.1145/3065386}
\BIBentrySTDinterwordspacing

\bibitem{simonyan2014very}
K.~Simonyan and A.~Zisserman, ``Very deep convolutional networks for large-scale image recognition,'' \emph{arXiv preprint arXiv:1409.1556}, 2014.

\bibitem{szegedy2015going}
C.~Szegedy, W.~Liu, Y.~Jia, P.~Sermanet, S.~Reed, D.~Anguelov, D.~Erhan, V.~Vanhoucke, and A.~Rabinovich, ``Going deeper with convolutions,'' in \emph{Proceedings of the IEEE conference on computer vision and pattern recognition}, 2015, pp. 1--9.

\bibitem{he2016deep}
K.~He, X.~Zhang, S.~Ren, and J.~Sun, ``Deep residual learning for image recognition,'' in \emph{Proceedings of the IEEE conference on computer vision and pattern recognition}, 2016, pp. 770--778.

\bibitem{Pak2017}
\BIBentryALTinterwordspacing
M.~Pak and S.~Kim, ``A review of deep learning in image recognition,'' in \emph{2017 4th International Conference on Computer Applications and Information Processing Technology ({CAIPT})}.\hskip 1em plus 0.5em minus 0.4em\relax {IEEE}, Aug. 2017. [Online]. Available: \url{https://doi.org/10.1109/caipt.2017.8320684}
\BIBentrySTDinterwordspacing

\bibitem{HABIB20224244}
\BIBentryALTinterwordspacing
G.~Habib and S.~Qureshi, ``Optimization and acceleration of convolutional neural networks: A survey,'' \emph{Journal of King Saud University - Computer and Information Sciences}, vol.~34, no.~7, pp. 4244--4268, 2022. [Online]. Available: \url{https://www.sciencedirect.com/science/article/pii/S1319157820304845}
\BIBentrySTDinterwordspacing

\bibitem{rcnn}
R.~Girshick, J.~Donahue, T.~Darrell, and J.~Malik, ``Rich feature hierarchies for accurate object detection and semantic segmentation,'' 2014.

\bibitem{fastrcnn}
R.~Girshick, ``Fast r-cnn,'' in \emph{Proceedings of the IEEE international conference on computer vision}, 2015, pp. 1440--1448.

\bibitem{fasterrcnn}
S.~Ren, K.~He, R.~Girshick, and J.~Sun, ``Faster r-cnn: Towards real-time object detection with region proposal networks,'' \emph{Advances in neural information processing systems}, vol.~28, 2015.

\bibitem{Faster_R-CNN}
\BIBentryALTinterwordspacing
------, ``Faster r-cnn: Towards real-time object detection with region proposal networks,'' 2015. [Online]. Available: \url{https://arxiv.org/abs/1506.01497}
\BIBentrySTDinterwordspacing

\bibitem{YOLO}
\BIBentryALTinterwordspacing
J.~Redmon, S.~Divvala, R.~Girshick, and A.~Farhadi, ``You only look once: Unified, real-time object detection,'' 2015. [Online]. Available: \url{https://arxiv.org/abs/1506.02640}
\BIBentrySTDinterwordspacing

\bibitem{Pengnoo_9121234}
M.~Pengnoo, M.~T. Barros, L.~Wuttisittikulkij, B.~Butler, A.~Davy, and S.~Balasubramaniam, ``Digital twin for metasurface reflector management in 6g terahertz communications,'' \emph{IEEE Access}, vol.~8, pp. 114\,580--114\,596, 2020.

\bibitem{YOLOv5}
\BIBentryALTinterwordspacing
G.~Jocher, ``Yolov5,'' 2020. [Online]. Available: \url{https://github.com/ultralytics/yolov5}
\BIBentrySTDinterwordspacing

\bibitem{sozzi2022automatic}
M.~Sozzi, S.~Cantalamessa, A.~Cogato, A.~Kayad, and F.~Marinello, ``Automatic bunch detection in white grape varieties using yolov3, yolov4, and yolov5 deep learning algorithms,'' \emph{Agronomy}, vol.~12, no.~2, p. 319, 2022.

\bibitem{Mukhopadhyay2019}
\BIBentryALTinterwordspacing
A.~Mukhopadhyay, I.~Mukherjee, and P.~Biswas, ``Comparing {CNNs} for non-conventional traffic participants,'' in \emph{Proceedings of the 11th International Conference on Automotive User Interfaces and Interactive Vehicular Applications: Adjunct Proceedings}.\hskip 1em plus 0.5em minus 0.4em\relax {ACM}, Sep. 2019. [Online]. Available: \url{https://doi.org/10.1145/3349263.3351336}
\BIBentrySTDinterwordspacing

\bibitem{mediapipe}
C.~Lugaresi, J.~Tang, H.~Nash, C.~McClanahan, E.~Uboweja, M.~Hays, F.~Zhang, C.-L. Chang, M.~G. Yong, J.~Lee \emph{et~al.}, ``Mediapipe: A framework for building perception pipelines,'' \emph{arXiv preprint arXiv:1906.08172}, 2019.

\bibitem{swin}
Z.~Liu, Y.~Lin, Y.~Cao, H.~Hu, Y.~Wei, Z.~Zhang, S.~Lin, and B.~Guo, ``Swin transformer: Hierarchical vision transformer using shifted windows,'' in \emph{Proceedings of the IEEE/CVF international conference on computer vision}, 2021, pp. 10\,012--10\,022.

\bibitem{Wang2021}
\BIBentryALTinterwordspacing
T.~Wang, J.~Li, Y.~Deng, C.~Wang, H.~Snoussi, and F.~Tao, ``Digital twin for human-machine interaction with convolutional neural network,'' \emph{International Journal of Computer Integrated Manufacturing}, vol.~34, no. 7-8, pp. 888--897, May 2021. [Online]. Available: \url{https://doi.org/10.1080/0951192x.2021.1925966}
\BIBentrySTDinterwordspacing

\bibitem{Ji2013-ay}
S.~Ji, W.~Xu, M.~Yang, and K.~Yu, ``{3D} convolutional neural networks for human action recognition,'' \emph{IEEE Trans. Pattern Anal. Mach. Intell.}, vol.~35, no.~1, pp. 221--231, Jan. 2013.

\bibitem{yolov3}
J.~Redmon and A.~Farhadi, ``Yolov3: An incremental improvement,'' \emph{arXiv preprint arXiv:1804.02767}, 2018.

\end{thebibliography}

\end{document}